\pdfoutput=1

\documentclass[11pt]{article}
\usepackage{naacl2021}
\usepackage{times}
\usepackage{latexsym}
\usepackage[T1]{fontenc}
\usepackage[utf8]{inputenc}
\usepackage{booktabs}
\usepackage{amsmath}
\usepackage{amssymb}
\usepackage{array}
\usepackage{inconsolata}
\usepackage{multirow}
\usepackage{graphicx}
\usepackage{enumitem}
\usepackage{multicol}

\usepackage{microtype}

\title{Fine-tuning Encoders for Improved Monolingual and\\
Zero-shot Polylingual Neural Topic Modeling}

\author{Aaron Mueller \and Mark Dredze \\
  Department of Computer Science \\
  Johns Hopkins University \\
  \texttt{amueller@jhu.edu, mdredze@cs.jhu.edu}
}

\date{}

\begin{document}
\maketitle

\begin{abstract}
Neural topic models can augment or replace bag-of-words inputs with the learned representations of deep pre-trained transformer-based word prediction models. One added benefit when using representations from multilingual models is that they facilitate zero-shot polylingual topic modeling. However, while it has been widely observed that pre-trained embeddings should be fine-tuned to a given task, it is not immediately clear what supervision should look like for an unsupervised task such as topic modeling. Thus, we propose several methods for fine-tuning encoders to improve both monolingual and zero-shot polylingual neural topic modeling. We consider fine-tuning on auxiliary tasks, constructing a new topic classification task, integrating the topic classification objective directly into topic model training, and continued pre-training. We find that fine-tuning encoder representations on topic classification and integrating the topic classification task directly into topic modeling improves topic quality, and that fine-tuning encoder representations on any task is the most important factor for facilitating cross-lingual transfer.
\end{abstract}

\section{Introduction}
Topic models \citep{lda} are widely used across numerous disciplines to study large corpora \cite{graber2017applications}. These data-driven models discover salient themes and semantic clusters without any supervision. Monolingual topic models are language-agnostic but do not align topics across languages, as they have a fixed language-specific vocabulary which cannot be aligned cross-lingually after training. Polylingual topic models \citep{mimn2009polylingual}, however, enable users to consider multilingual corpora, and to discover and align topics across languages.

Recent work has demonstrated the effectiveness of deep transformer-based language models to encode text documents for a wide variety of applications \cite{xia2020bert}. Furthermore, when trained on multilingual corpora, they have been able to discover cross-lingual alignments despite the lack of explicit cross-lingual links \cite{wu2019beto}.
Models such as multilingual BERT (mBERT; \citealp{bert}) or XLM-RoBERTa (XLM-R; \citealp{xlmr}) can produce a representation of text in a shared subspace across multiple input languages, suitable for both monolingual and multilingual settings, including zero-shot language transfer \citep{pires2019zeroshot}.

Simultaneously, topic models have increasingly incorporated neural components. This has included inference networks which learn representations of the input document \cite{miao2017discovering,neuralprodlda} that improve over using bags of words directly, as well as replacing bags of words with contextual representations. In particular, the latter allows topic models to benefit from pre-training on large corpora. 
For example, contextualized topic models (CTMs) \cite{ctm_1} use autoencoded contextual sentence representations of input documents.

An intriguing advantage of using encoders in topic models is their latent multilinguality. 
Polylingual topic models \cite{mimn2009polylingual} are lightweight in their cross-lingual supervision to align topics across languages, but they nonetheless require some form of cross-lingual alignment. While the diversity of resources and approaches for training polylingual topic models enable us to consider many language pairs and domains, there may be cases where existing resources cannot support an intended use case. Can topic models become polylingual models by relying on multilingual encoders even without additional alignments? 

\citet{ctm_1} show that CTMs based on contextual sentence representations enable zero-shot cross-lingual topic transfer. While promising, this line of work omits a key step in using contextualized embeddings: fine-tuning.
It has been widely observed that task specific fine-tuning of pretrained embeddings, even with a small amount of supervised data, can significantly improve performance on many tasks, including in zero- and few-shot settings \citep{howard2018universal,wu2019beto}. 
However, in the case of unsupervised topic modeling, from where are we to obtain task-specific supervised training data?

We propose an investigation of how supervision should be bootstrapped to improve language encoders for monolingual and polylingual topic model learning. We also propose a set of experiments to better understand \emph{why} certain forms of supervision are effective in this unsupervised task. Our contributions include the following:
\begin{enumerate}[noitemsep,topsep=0pt,partopsep=0pt]
    \item We fine-tune contextualized sentence embeddings on various established auxiliary tasks, finding that many different tasks can be used to improve downstream topic quality and zero-shot topic model transfer.
    \item We construct fine-tuning supervision for sentence embeddings through a proposed topic classification task, showing further improved topic coherence. This task uses only the data on which we perform topic modeling.
    \item We integrate a topic classification objective directly into the neural topic model architecture (without fine-tuning the embeddings) to understand whether the embeddings or the topic classification objective is responsible for performance improvements. We find that this approach improves topic quality but has little effect on cross-language topic transfer.
\end{enumerate}
We present results for both monolingual topic models and cross-lingual topic transfer from English to French, German, Portuguese, and Dutch.

Our code, including instructions for replicating our dataset and experimental setup, are publicly available on GitHub.\footnote{\url{https://github.com/aaronmueller/contextualized-topic-models}}

%\section{Related Work}
%\subsection{Multilingual Topic Modeling}

%\subsection{Neural Topic Modeling}

\section{Background}
\paragraph{Neural Topic Models}
Neural topic models (NTMs) are defined by their parameterization by (deep) neural networks or incorporation of neural elements. This approach has become practical largely due to advances in variational inference---specifically, variational autoencoders (VAEs; \citealp{vae}). The Neural Variational Document Model \citep{miao2016neural} and Gaussian Softmax Model \citep{miao2017discovering} rely on amortized variational inference to approximate the posterior \citep{zhao2017infovae,krishnan2018challenges}. As these methods employ Gaussian priors, they use softmax transforms to ensure non-negative samples. Another approach has used ReLU transforms \citep{ding2018coherenceaware}.

Conversely, ProdLDA \citep{neuralprodlda} uses a Dirichlet prior that produces non-negative samples which do not need to be transformed. ProdLDA uses an inference network with a VAE to map from an input bag of words to a continuous latent representation. The decoder network samples from this hidden representation to form latent topic representations. Bags of words are reconstructed for each latent space; these constitute the output topics. Others have reported that ProdLDA is the best-performing NTM with respect to topic coherence \cite{miao2017discovering}.

Contextualized topic models (CTMs; \citealp{ctm_1,ctm_2}) extend ProdLDA by replacing the input bag of words with sentence-BERT (SBERT; \citealp{sbert}) embeddings. If the SBERT embeddings are based on a multilingual model such as mBERT \citep{bert} or XLM-R \citep{xlmr}, then the topic model becomes implicitly polylingual due to the unsupervised alignments induced between languages during pre-training. This is distinct from how polylinguality is induced in approaches based on Latent Dirichlet Allocation (LDA; \citealp{lda}), which require some form of cross-lingual alignments \cite{mimn2009polylingual}.

Using embeddings in topic models is not new \citep{das2015gaussian,liu2015topical,li2016auxembed}. While a few recent approaches have leveraged word embeddings for topic modeling \citep{gupta2019docneural,dieng2020embed,sia2020clusters}, none of these have investigated cross-lingual topic transfer.

\paragraph{Polylingual Topic Models}
Polylingual topic models require some form of cross-lingual alignments, which can come from comparable documents \citep{mimn2009polylingual}, word alignments \citep{polyling_wordalign}, multilingual dictionaries \citep{polyling_dict}, code-switched documents \citep{polyling_codeswitch}, or other distant alignments such as anchors \citep{polyling_anchor}. Work on incomparable documents with soft document links \citep{polyling_incomparable} still relies on dictionaries.

While these types of alignments have been common in multilingual learning \cite{ruder2019survey}, they no longer represent the state-of-the-art. More recent approaches instead tend to employ large pre-trained multilingual models \citep{wu2019beto} that induce unsupervised alignments between languages during pre-training.

\section{Fine-Tuning Encoders}
Fine-tuning is known to improve an encoder's representations for a specific task when data directly related to the task is present \citep{howard2018universal,wu2019beto}. Nonetheless, this requires supervised data, which is absent in unsupervised tasks like ours. We consider several approaches to create fine-tuning supervision for topic modeling.

\subsection{Fine-tuning on Related Tasks}
In the absence of supervised training sets, transfer learning can be used to learn from one supervised task (or many tasks in the case of meta-learning) for improvements on another \citep{ruder2019transfer}. While transfer is typically performed from a pre-trained masked language model to downstream fine-tuning tasks, transfer can also be performed from one fine-tuning task to another. The aim is that the auxiliary task should induce representations similar to those needed for the target task.

What task can serve as an effective auxiliary task for topic modeling? We turn to document classification, the task of identifying the primary topic present in a document from a fixed set of (typically human-identified and human-labeled) topics. We may not have a document classification dataset from the same domain as the topic modeling corpus, nor a dataset which uses the same topics as those present in the corpus. However, fine-tuning could teach the encoder to produce topic-level document representations, regardless of the specific topics present in the data. We use MLDoc \citep{mldoc}, a multilingual news document classification dataset and fine-tune on English.

For comparison, we fine-tune on a natural language inference (NLI) task. While it is not closely related to topic modeling, this task is a popular choice for fine-tuning both word and sentence representations. This allows us to measure how much task relatedness matters for fine-tuning. 

\subsection{Fine-tuning on Topic Models}
The auxiliary tasks use data from a different domain (and task) than the domain of interest for the topic model. Can we bootstrap more direct supervision on our data?

We employ an LDA-based topic model to produce a form of topic supervision. We first run LDA on the target corpus to generate topic distributions for each document. Then, we use the inferred topic distributions as supervision by labeling each document with its most probable topic.
We fine-tune on this data as we did for the document classification task; the setup is identical except for how the labels are obtained. The advantage of this method is that LDA topics can be created for any corpus.

% Self training paper (FB) says that continued pre-training doesn't work

\subsection{Continued Pre-Training}
\citet{dontstoppretraining} advocated for adapting an encoder to the domain on which one will later fine-tune. This is done by performing continued pre-training over in-domain data using the masked language modeling (MLM) objective.\footnote{We note that there are mixed findings in the literature with respect to this method \cite{du2020self}.} Because continued pre-training requires no task-specific supervision, and because topic modeling implies a sizeable corpus of in-domain documents, we consider continued pre-training on the target corpus as another approach to adapting an encoder. As continued pre-training can be done before fine-tuning, we also try doing both.

Does topic classification improve performance because fine-tuning itself induces better representations for topic modeling, or because the model has been exposed to in-domain data and/or supervision directly from the target corpus before topic modeling? Continued pre-training on the target corpus may allow us to answer this question, and provides a further approach for adapting encoders to specific domains.

\subsection{Modifying the Topic Modeling Objective}\label{sec:tcctm}
\begin{figure}[t]
    \centering
    \includegraphics[width=\columnwidth]{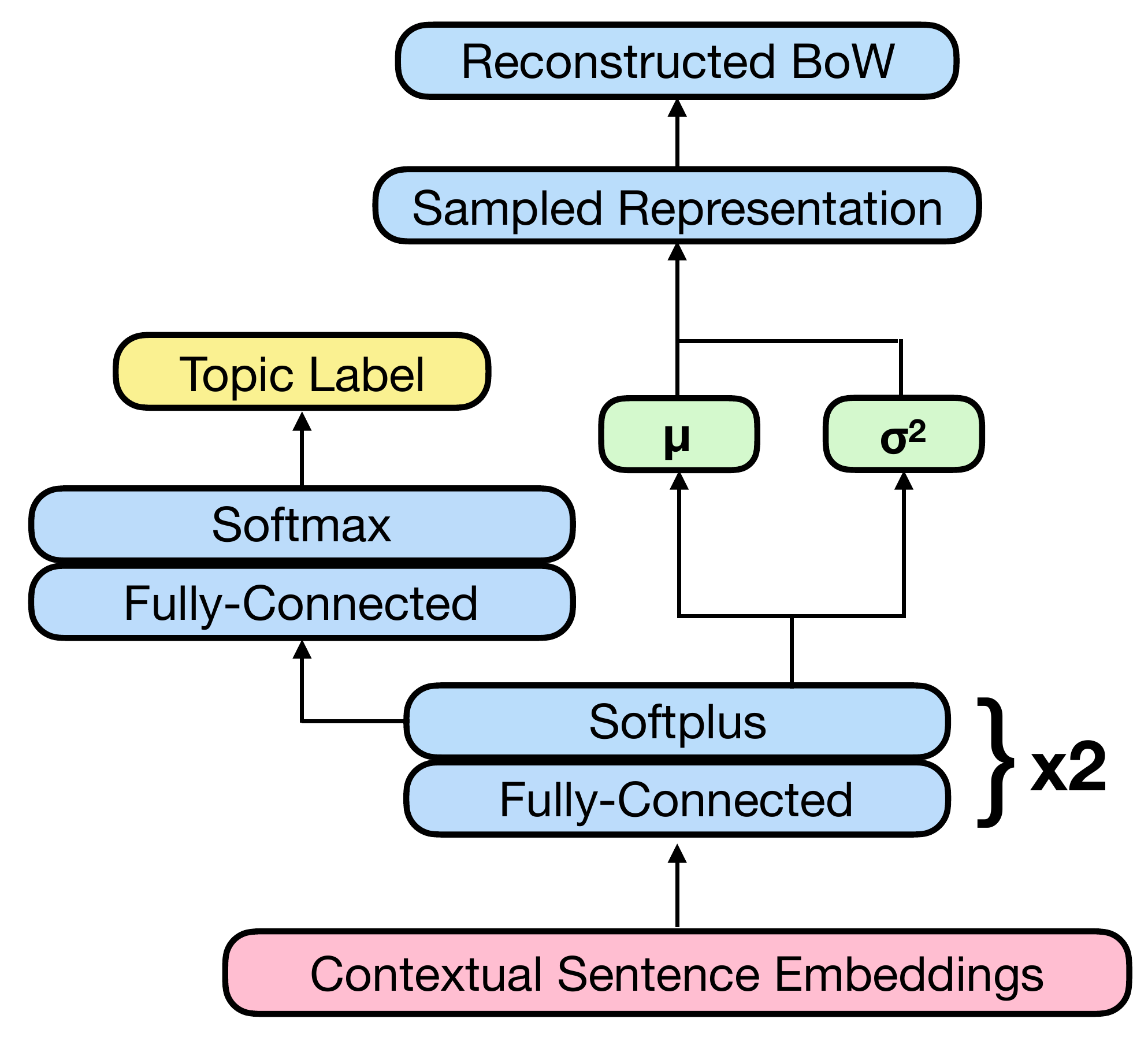}
    \caption{Architecture used in the topic classification contextualized topic model (TCCTM) approach (\S\ref{sec:tcctm}). This is similar to the architecture of \citet{ctm_1}, but with an added fully-connected layer and softmax to produce a topic classification for the input document from its hidden representation.}
    \label{fig:new_arch}
\end{figure}

Both continued pre-training and fine-tuning provide supervision for our target task, but both create dependence on a pipeline: we must train and/or fine-tune sentence embeddings, then train a neural topic model using the modified embeddings.

However, we can combine the topic classification task and topic modeling into a single end-to-end procedure by modifying the inference network of the CTM. Figure~\ref{fig:new_arch} shows our proposed architecture: a fully-connected layer into a softmax to predict the topic label of the document based on the learned representation of the VAE. Note that we do not necessarily expect this architecture to outperform fine-tuning sentence embeddings: rather, this architecture allows us to ablate over the location of the topic classification objective, which allows us to determine whether improvements in topic quality and/or transfer are due to improved sentence embeddings induced by fine-tuning, or due to the topic classification task itself.

We use the negative log-likelihood loss between the topic predicted by LDA (which we treat as the true label) and the topic predicted by our model, adding this loss term (weighted by a hyperparameter $\lambda$) to the contextualized topic model's loss function. Thus, the new loss becomes
    \begin{align*}
    \mathcal{L}_{\text{TCCTM}} = \mathcal{L}_{\text{ELBO}} + \lambda \mathcal{L}_{\text{NLL}}
    \end{align*}
where $\mathcal{L}_{\text{ELBO}}$ is the negated evidence lower bound objective of the CTM, and $\mathcal{L}_{\text{NLL}}$ is the negative log-likelihood loss over topic classifications. We refer to this as the \emph{topic classification contextualized topic modeling} (TCCTM) loss, denoted $\mathcal{L}_{\text{TCCTM}}$.

TCCTM modifies the topic model, but not the embeddings. This approach is therefore orthogonal to fine-tuning, and the two approaches can be combined; thus, we test the performance of TCCTM with and without fine-tuning.

% When encoding the embeddings, we add an additional dimension which corresponds to the label of the document. Thus, for any input document, the first layer of the VAE must minimize both reconstruction error and correctly predict the label of the document. The second layer then uses the predicted label and encoded document to form the hidden representation of the document, which acts as input for the decoder network. This is similar to conditional variational autoencoders (CVAEs), but rather than treating the document label as a latent variable, we explicitly minimize cross-entropy loss over document labels.

\section{Experiments}
% any stats about dimensionality, learning rate, grid search, etc. See the list here: https://arxiv.org/abs/1909.03004
% This paragraph should be added in this section:
% We try TCCTM based on sentence embeddings from an un-tuned sentence-transformer model, as well as models fine-tuned on document classification or NLI. We do not perform this approach on a model fine-tuned on topic classification to avoid overfitting and confounds from performing the same task in multiple stages of the model.

\paragraph{Data} We begin by creating a multilingual dataset for topic modeling based on aligned Wikipedia articles extracted from Wikipedia Comparable Corpora\footnote{\url{https://linguatools.org/tools/corpora/wikipedia-comparable-corpora/}} in English, French, German, Portuguese, and Dutch. We use 100,000 English articles for training the topic models and evaluating monolingual topic coherence. We also extract 100,000 aligned articles for each language to build comparable vocabularies for preprocessing the test data.\footnote{We release article IDs and splits with our code.} For each language, we use a vocabulary of the 5,000 most frequent word types (case-insensitive), excluding stopwords---25,000 types total. We use the English training articles to evaluate monolingual topic quality, and hold out for cross-lingual evaluation a set of 10,000 aligned test articles per-language.

For out-of-domain topic classification, we use a dataset of COVID academic articles (in English).\footnote{\url{https://www.kaggle.com/allen-institute-for-ai/CORD-19-research-challenge}} To facilitate comparison with the Wikipedia dataset, we extract 100,000 articles and use a vocabulary size of 5,000.

To obtain topic labels for each English document, we run LDA for $400$ iterations and choose the number of topics $\tau$ by performing a search in $\{10, 20,\ldots,250\}$, optimizing over NPMI coherence. We find that $\tau \in \{100, 110, 120\}$ is best and use $\tau=100$ here. We label each document with its most probable topic by counting the number of tokens in the document in the top-$10$ token list for each topic, then taking the argmax. We perform the same procedure on the out-of-domain COVID dataset to generate out-of-domain topic classification supervision, finding that $\tau=80$ is best on this dataset with respect to NPMI coherence.

For the document classification task, we use MLDoc \citep{mldoc}, a multilingual news dataset; we fine-tune on the English data. For NLI, we follow \citet{reimers2020multilingsbert} in using a mixture of SNLI \citep{snli} and MultiNLI \citep{multinli}, both of which only contain English data.

\paragraph{Training Details}
We consider embeddings produced by both  mBERT \citep{bert} and XLM-R \citep{xlmr}.
For fine-tuning, we append to these models a fully-connected layer followed by a softmax, using a negative log-likelihood loss for topic/document classification. We perform a search over the number of epochs in the range $[1,8]$, optimizing over downstream NPMI coherence during topic modeling.

We follow the procedure of \citet{sbert} to create sentence embedding models from contextual word representations: we mean-pool word embeddings for two sentences simultaneously, feeding these as inputs to a softmax classifier. We use batch size $16$; other hyperparameters are kept from \citet{sbert}. 
%We also try pooling the embeddings of mBERT/XLM-R without fine-tuning.

For NLI fine-tuning, we follow the procedure and use the hyperparameters of \citet{reimers2020multilingsbert}: we first fine-tune monolingual BERT on SNLI and MultiNLI; the embeddings are pooled during fine-tuning to create a sentence embedding model. We then perform a knowledge distillation step from the monolingual SBERT model to XLM-R or mBERT.

Continued pre-training is performed by training with the MLM objective on English Wikipedia. We run for $1$ epoch, using gradient accumulation to achieve an effective batch size of $256$. We can pool the embeddings from the resulting model directly or perform fine-tuning after continued pre-training.

\begin{table*}[t]
    \centering
    \resizebox{\linewidth}{!}{
    \begin{tabular}{llrr|p{1cm}p{1.9cm}p{2cm}p{1.7cm}}
        \toprule
        \bf{Model} & \bf{Fine-tuning} & \multicolumn{2}{r}{\bf{NPMI}} & Neural model & Fine-tuned embeddings & Topic\newline classification & In-domain data \\
        \midrule
        LDA & -- & & 0.129 & & & & \\
        ProdLDA & -- & & 0.129 & \centering\checkmark & & & \\
        \midrule
        \multirow{6}{*}{CTM} & & XLM-R & mBERT  & & & & \\\cmidrule(lr){3-4}
            & None & 0.144 & 0.144 & \centering\checkmark & & &\\
            & NLI & 0.153 & 0.152 & \centering\checkmark & \centering\checkmark & &\\
            & Doc.\ Class. & 0.156 & 0.153 & \centering\checkmark & \centering\checkmark & & \\
            & Topic Class.\ (COVID) & 0.156 & 0.153 & \centering\checkmark & \centering\checkmark & \centering\checkmark & \\
            & Topic Class.\ (Wiki) & \bf{0.160} & \bf{0.156} & \centering\checkmark & \centering\checkmark & \centering\checkmark & \centering\checkmark \tabularnewline
        \midrule
        \multirow{4}{*}{CPT+CTM} & None & 0.147 & 0.147 & \centering\checkmark & & & \centering\checkmark \tabularnewline
        & NLI & 0.150 & 0.149 & \centering\checkmark & \centering\checkmark & & \centering\checkmark \tabularnewline
        & Topic Class.\ (COVID) & 0.148 & 0.147 & \centering\checkmark & \centering\checkmark & \centering\checkmark & \centering\checkmark \tabularnewline
        & Topic Class.\ (Wiki) & 0.151 & 0.149 & \centering\checkmark & \centering\checkmark & \centering\checkmark & \centering\checkmark \tabularnewline
        \midrule
        \multirow{3}{*}{TCCTM} & None & 0.157 & 0.154 & \centering\checkmark & & \centering\checkmark & \\
            & NLI & 0.152 & 0.151 & \centering\checkmark & \centering\checkmark & \centering\checkmark & \\
            & Doc.\ Class. & 0.153 & 0.152 & \centering\checkmark & \centering\checkmark & \centering\checkmark & \\
        \bottomrule
    \end{tabular}}
    \caption{NPMI coherences for contextualized topic models (CTM), CTMs with continued pre-training (CPT+CTM), and the TCCTM model on the English Wikipedia dataset. We present results with and without fine-tuning for XLM-R and mBERT-based sentence embeddings. The right side of the table indicates whether each setup is based on a neural architecture, whether the SBERT embeddings are fine-tuned before topic modeling, whether the topic classification task/objective is present, and whether the embeddings have been trained/fine-tuned on the same data later used for topic modeling.}
    \label{tab:monoling}
\end{table*}

%\begin{table}[t]
%    \centering
%    \resizebox{\linewidth}{!}{
%    \begin{tabular}{llrr}
%    \toprule
%        & & \multicolumn{2}{c}{\bf{NPMI}} \\\cmidrule(lr){3-4}
%        \bf{Fine-tuning Task} & \bf{Data} & XLM-R & mBERT \\
%        \midrule
%        No fine-tuning & --- & 0.144 & 0.144 \\
%        \midrule
%        NLI & SNLI+MultiNLI & 0.153 & 0.152 \\
%        % Sentence Encoding & & 0.149 & 0.147 \\
%        \midrule
%        Doc.\ class. & MLDoc & 0.156 & 0.153 \\
%        %\multirow{2}{*}{Topic class.} & IQOS & 0.155 & 0.150 \\
%        Topic class. & Patents & 0.154 & 0.153 \\
%        Topic class. & Wiki & \bf{0.160} & 0.156 \\
%        \bottomrule
%    \end{tabular}}
%    \caption{NPMI coherences for contextualized topic models ($\tau=100$) on the English Wikipedia dataset after fine-tuning encoders on various tasks and datasets.}
%    \label{tab:doc-class}
%\end{table}

When topic modeling, we run the CTM for $60$ epochs, using an initial learning rate of $2\times 10^{-3}$, dropout $0.2$, and batch size $64$. The VAE consists of two hidden layers of dimensionality $100$ (as in \citealt{neuralprodlda} and \citealt{ctm_2}). The ProdLDA baseline is run using the same hyperparameters and the same architecture as a CTM, differing only in using bags of words as input instead of SBERT representations.

For the LDA baseline, we employ MalletLDA \citep{mallet} as implemented in the \texttt{gensim} wrapper, running for $400$ iterations on the Wikipedia data using $\tau=100$.

We fine-tune $\lambda$ in the TCCTM objective in $\{0.1,0.2,\ldots,3.0\}$, finding that $\lambda=1.0$ yields the best downstream topic coherence for the target Wikipedia data. We try TCCTM based on non-fine-tuned sentence embeddings, as well as models fine-tuned on document classification or NLI. We do not perform this approach on a model fine-tuned on in-domain topic classification to avoid overfitting and confounds from performing the same task in multiple stages of the model.

\paragraph{Evaluation} 

To evaluate topic quality, we measure normalized pointwise mutual information (NPMI) coherence on the English Wikipedia dataset. NPMI is used because it is comparable across architectures and objectives, and because it tends to correlate better with human judgments of topic quality \citep{npmi}. While perplexity has been used to evaluate LDA \citep{lda} as well as neural topic models in the past \citep{miao2017discovering}, it is not comparable across different objective functions when using neural approaches (as it depends on the test loss) and tends to correlate poorly with human judgments \citep{chang2009tealeaves}. Topic significance ranking \citep{alsumait09tsr} has been used to measure and rank topics by semantic importance/relevance, though we care more about overall topic quality than ranking topics.

As the contextualized topic model is based on a multilingual encoder, it is able to generate $\theta_i$ (a topic distribution over document $i$) given input embeddings from a document $\mathbf{h}_i$ in any language it has seen. To evaluate multilingual generalization, we measure the proportion of aligned test documents for which the most probable English topic $\theta_{i_{\text{English}}}$ is the same as the most probable target-language topic $\theta_{i_{\text{Target}}}$ (the \underline{Match} metric). We also measure the KL divergence between topic distributions $\text{D}_\text{KL}(\theta_{i_{\text{English}}} \Vert \theta_{i_{\text{Target}}})$, taking the mean over all aligned documents (the \underline{KL} metric). We construct a random baseline by randomly shuffling the English articles and then computing both metrics against the newly unaligned foreign articles.
    %\begin{align*}
    %    \text{Match} = \frac{\sum_{\theta_i \in \Theta} \arg\max\theta_{i_{\text{English}}} = \arg\max\theta_{i_{\text{Target}}}}{|\Theta|}
    %\end{align*}

\begin{table*}[ht]
    \centering
    \resizebox{\linewidth}{!}{
    \begin{tabular}{lrrrrrrrrrr}
    \toprule
       & \multicolumn{2}{c}{French} & \multicolumn{2}{c}{German} & \multicolumn{2}{c}{Portuguese} & \multicolumn{2}{c}{Dutch} & \multicolumn{2}{c}{\textsc{Mean}} \\\cmidrule(lr){2-3}\cmidrule(lr){4-5}\cmidrule(lr){6-7}\cmidrule{8-9}\cmidrule(lr){10-11}
    Model & Match & KL & Match & KL & Match & KL & Match & KL & Match & KL \\
    \midrule
    CTM (No FT) & 20.11 & 0.71 & 41.68 & 0.46 & 24.85 & 0.67 & 46.74 & 0.40 & 33.30 & 0.56 \\
    CTM+FT (NLI) & \bf{53.68} & 0.39 & \bf{56.29} & \bf{0.33} & \bf{54.38} & \bf{0.36} & \bf{56.98} & 0.31 & \bf{55.33} & 0.35 \\
    CTM+FT (DC) & 35.53 & 0.61 & 42.09 & 0.49 & 38.12 & 0.53 & 49.70 & 0.40 & 41.36 & 0.51 \\
    CTM+FT (TC, COVID) & 41.09 & 0.54 & 46.39 & 0.47 & 43.56 & 0.48 & 51.11 & 0.40 & 45.54 & 0.47 \\
    CTM+FT (TC, Wiki) & 45.02 & 0.50 & 51.11 & 0.40 & 42.58 & 0.49 & 50.68 & 0.40 & 47.17 & 0.44 \\
    \midrule
    CPT (No FT) & 23.62 & 0.68 & 40.75 & 0.45 & 22.89 & 0.65 & 45.13 & 0.42 & 33.10 & 0.55 \\
    CPT+FT (NLI) & 43.43 & 0.45 &  48.09 & 0.38 & 43.04 & 0.46 & 49.53 & 0.38 & 46.02 & 0.42\\
    CPT+FT (TC, COVID) & 41.70 & 0.53 & 43.67 & 0.44 & 39.91 & 0.60 & 47.44 & 0.43 & 43.18 & 0.50\\
    CPT+FT (TC, Wiki) & 47.02 & 0.45 & 51.53 & 0.36 & 45.83 & 0.44 & 52.54 & 0.34 & 49.23 & 0.40 \\
    \midrule
    TCCTM (No FT) & 18.81 & 0.71 & 41.18 & 0.46 & 19.21 & 0.72 & 45.49 & 0.42 & 31.17 & 0.58 \\
    TCCTM+FT (NLI) & 53.30 & \bf{0.38} & 55.52 & \bf{0.33} & 53.75 & 0.37 & 56.40 & \bf{0.30} & 54.74 & \bf{0.34} \\
    TCCTM+FT (DC) & 41.83 & 0.51 & 48.72 & 0.42 & 38.80 & 0.53 & 49.73 & 0.39 & 44.77 & 0.46 \\
    \midrule
    Random & 0.92 & 1.48 & 1.22 & 1.39 & 1.24 & 1.48 & 1.09 & 1.44 & 1.12 & 1.44 \\
    \bottomrule
    \end{tabular}}
    \caption{Percentage of held-out documents assigned the same topic in English and other languages (Match, higher is better) and the mean KL divergence between the English and target language topic distributions per-document (KL, lower is better). For the random baseline, we compare randomly sampled English articles rather than using aligned articles.}
    \label{tab:multilingual}
\end{table*}

\section{Results}
\subsection{Monolingual Topic Modeling}
We compare topic coherences on the 100,000 English Wikipedia articles for LDA and ProdLDA baselines, a CTM with no fine-tuning, a CTM with continued pre-training (CPT), and the integrated TCCTM model. We also compare the effect of fine-tuning (FT) on the NLI task, on a document classification task (MLDoc), and on labels from LDA for the out-of-domain COVID dataset and for the in-domain Wikipedia data (Table~\ref{tab:monoling}). The baseline LDA and ProdLDA models both achieve the same coherence score of 0.129. Compared to these baselines, {\bf models based on contextualized representations always achieve higher topic coherence.}

We find that when using a base CTM without modifying its objective, {\bf fine-tuning on \emph{any} auxiliary task improves topic quality for CTMs}. Specifically, fine-tuning on in-domain topic classification data is best for monolingual topic modeling, followed closely by document classification on MLDoc. Topic classification on the out-of-domain COVID data results in the same topic coherence scores as document classification, indicating that topic classification is an effective method for bootstrapping supervision, even compared to established document classification datasets with human-labeled documents. The further gains in topic coherence when fine-tuning on Wikipedia topic classification data may be due to the data being in-domain, rather than due to the topic classification task. Fine-tuning on NLI yields less coherent topics than document or topic classification. For any given approach, XLM-R always outperforms mBERT.

We find that CPT without fine-tuning performs \emph{worse} than simply fine-tuning, but better than a CTM using embeddings which are not fine-tuned. Fine-tuning after performing continued pre-training (CPT+FT) slightly improves NPMI over CPT alone, but still results in less coherent topics than if we simply fine-tune on the in-domain Wiki data or the out-of-domain COVID data. Thus, the MLM objective seems to induce representations not conducive to topic modeling. Indeed, {\bf fine-tuning on any task is better than continuing to train the encoder on the exact data later used for the CTM.} This means that we may not attribute the effectiveness of topic classification solely to the model's seeing in-domain data before topic modeling; rather, some property of fine-tuning itself is better at inducing representations conducive to topic modeling.

Conversely, the TCCTM approach using non-fine-tuned embeddings produces more coherent topics than \emph{all} fine-tuning tasks except topic classification on in-domain Wikipedia data. This means that {\bf the topic classification task itself is also responsible for the high topic coherences observed}, and not just the fine-tuned sentence embeddings. Nonetheless, topic classification is more effective when used to fine-tune sentence embeddings, rather than as a part of the CTM objective---further cementing the importance of embeddings to topic quality. There seems to be interference---or perhaps overfitting---when combining TCCTM with embeddings fine-tuned on other tasks. Indeed, fine-tuning on document classification and NLI results in slightly less coherent topics than simply using TCCTM on non-fine-tuned sentence embeddings. Perhaps this could be mitigated with task-specific fine-tuning over $\lambda$; we leave this to future work.

\subsection{Polylingual Topic Modeling}

Table~\ref{tab:multilingual} presents results for zero-shot cross-lingual topic transfer. {\bf All models, including without fine-tuning, are far better than random chance on both metrics.} This indicates that multilingual encoders contain enough cross-lingual alignment as-is to induce cross-lingual topic alignment. Nonetheless, we also find that {\bf fine-tuning the embeddings on \emph{any} task produces better multilingual topic alignments than not fine-tuning}; NLI consistently shows the best cross-lingual transfer. Document classification is generally a worse fine-tuning task than topic classification for cross-lingual transfer, despite achieving similar monolingual performance.

\begin{figure}[t]
    \centering
    \includegraphics[width=\linewidth]{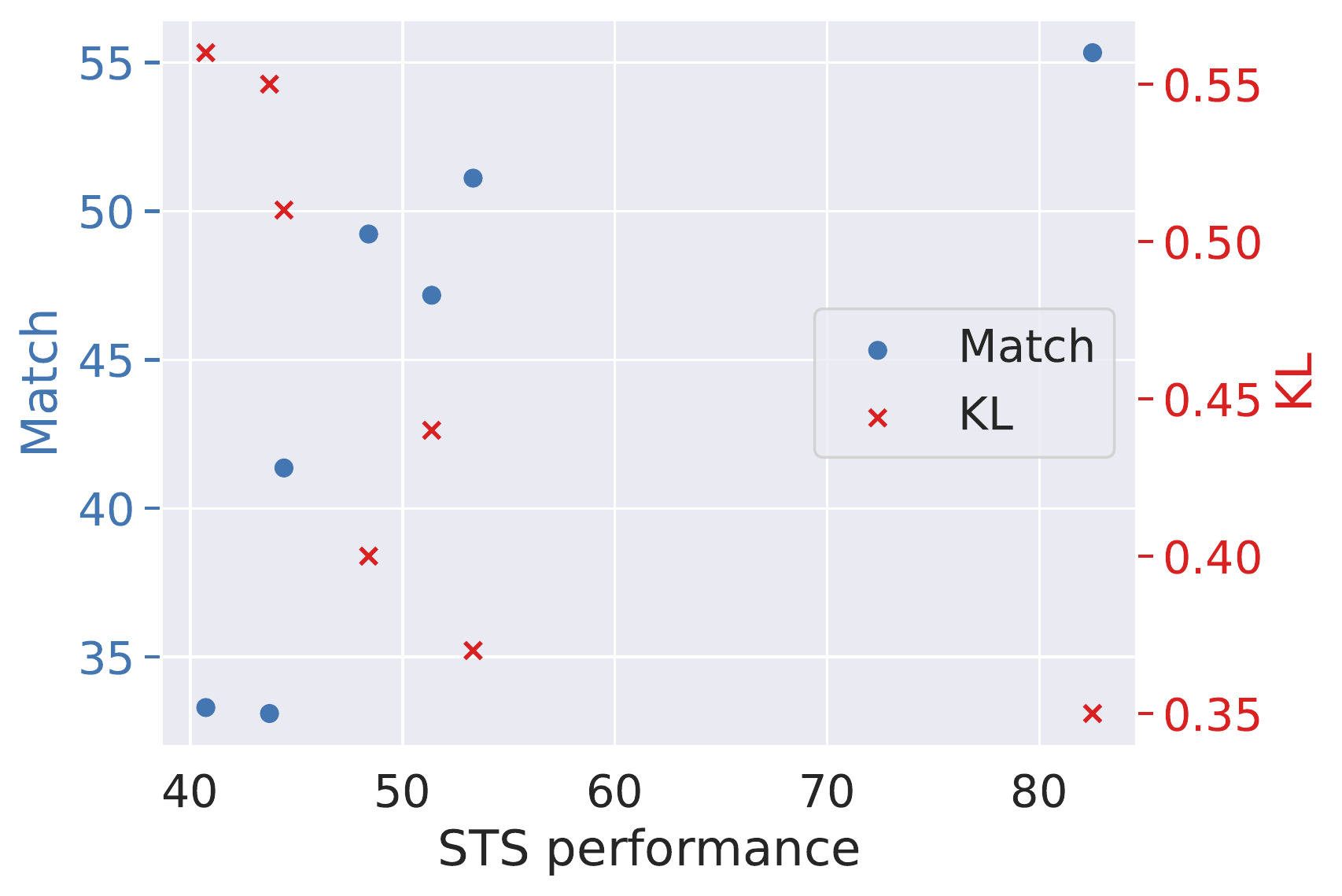}
    \caption{Performance on the Semantic Textual Similarity (STS) benchmark (Spearman correlation between cosine similarity of sentence representations and gold labels for STS tasks) versus mean Match and KL per-language for sentence embedding models fine-tuned on various tasks (all with XLM-R-based sentence embeddings.) The outlier is the XLM-R model fine-tuned on NLI, as it was explicitly designed and tuned for STS \citep{reimers2020multilingsbert}. We do not include TCCTM as it does not modify sentence embeddings.}
    \label{fig:sts}
\end{figure}

\begin{figure*}[ht]
    \centering
    \begin{minipage}{0.49\linewidth}
        \centering
        \includegraphics[width=\linewidth]{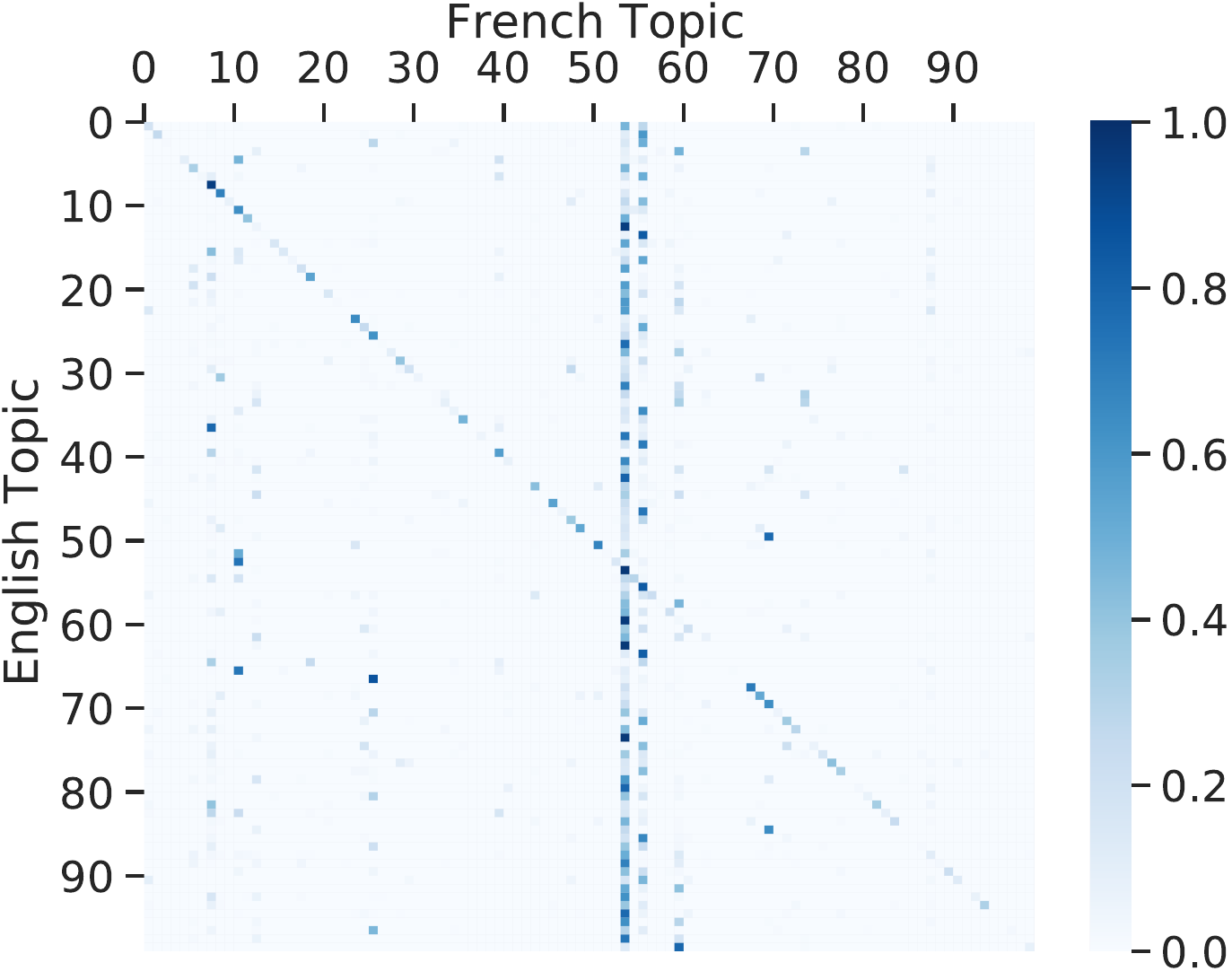}
    \end{minipage}
    \hfill
    \begin{minipage}{0.49\linewidth}
        \centering
        \includegraphics[width=\linewidth]{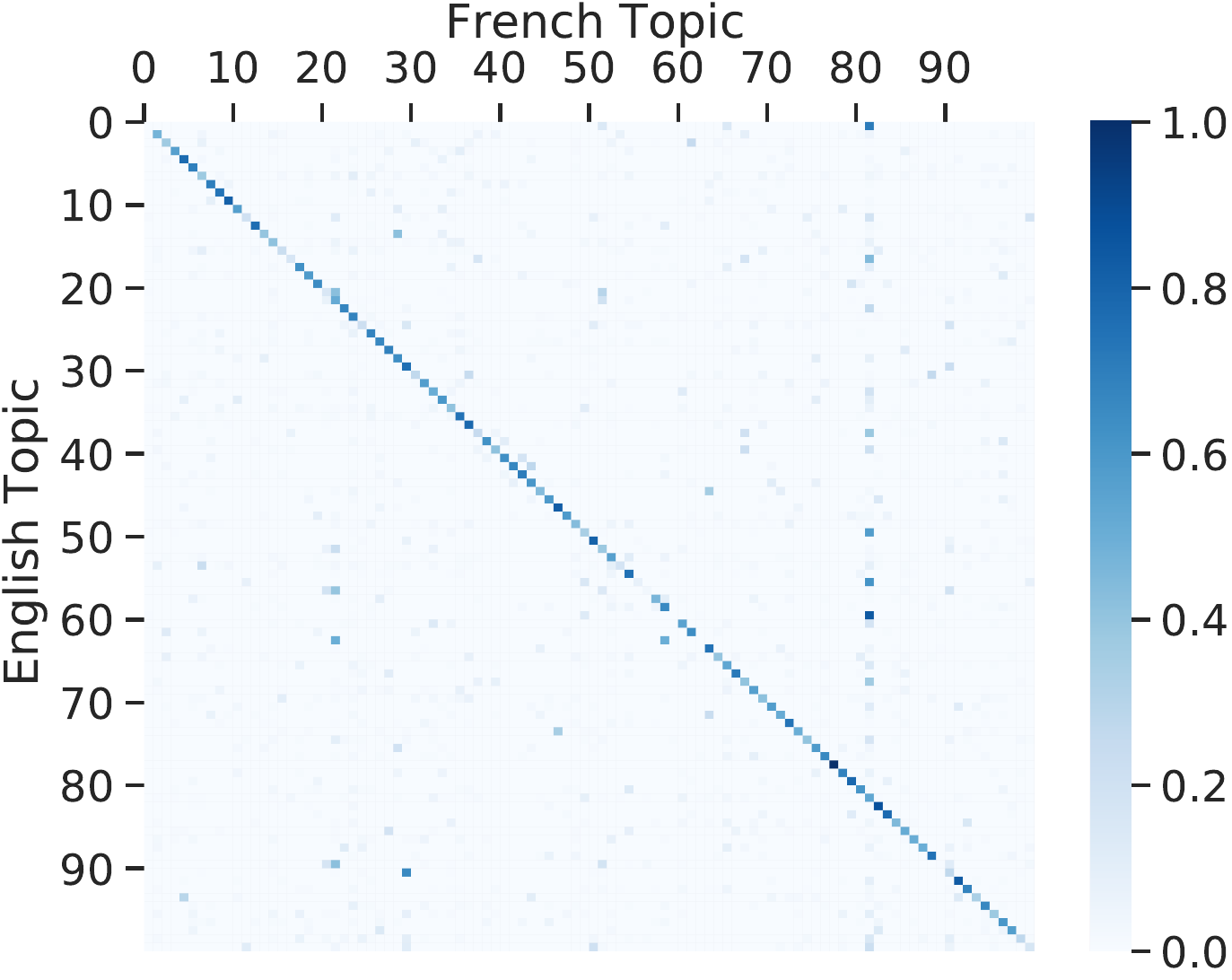}
    \end{minipage}
    \caption{Row-normalized confusion matrices comparing topic assignments from the contextualized topic model in English and French on aligned documents, both without fine-tuned sentence embeddings (left) and with embeddings fine-tuned on NLI (right). Both are based on XLM-R.}
    \label{fig:confmat}
\end{figure*}

When performing continued pre-training without fine-tuning, we find that results tend to be comparable to the CTM without fine-tuning, though slightly better. When performing both continued pre-training and fine-tuning, we achieve only slightly higher results compared to simply fine-tuning; thus, in both monolingual and multilingual settings, {\em the fine-tuning task is more important for topic transfer than seeing in-domain data or having a better in-domain language model.}

The TCCTM objective alone produces fairly poor multilingual topic alignments, despite its positive effect in monolingual contexts; however, it consistently performs effective cross-lingual transfer when paired with sentence embeddings fine-tuned on document classification. When paired with embeddings fine-tuned on NLI, TCCTM achieves almost the same scores as the CTM model using the same embeddings. Thus, {\bf the fine-tuning task used for the sentence embeddings is the most important factor for cross-lingual transfer}.

\begin{table*}[ht]
    \centering
    \resizebox{\linewidth}{!}{
    \begin{tabular}{lll}
    \toprule
    \bf{Lang} & \bf{Sample Document} & \bf{Topic} \\  
    \midrule
    \texttt{en} & Niccolò Zucchi was an Italian Jesuit, astronomer, and physicist\ldots & 12: star, constellation, sky, cluster, galaxy \\
    \texttt{fr} & Niccolò Zucchi\ldots était un prêtre jésuite italien, astronome et physicien\ldots & 12: star, constellation, sky, cluster, galaxy \\
    \texttt{pt} & Niccolò Zucchi foi um jesuíta, astrônomo e físico italiano\ldots & 12: star, constellation, sky, cluster, galaxy \\
    \texttt{de} & Niccolò Zucchius, auch Niccolo Zucchi, war ein italienischer Astronom und Physiker\ldots  & 12: star, constellation, sky, cluster, galaxy \\
    \texttt{nl} & Niccolò Zucchi was een Italiaans astronoom\ldots & 12: star, constellation, sky, cluster, galaxy \\
    \midrule
    \texttt{en} & Chambilly is a commune in the Saône-et-Loire department\ldots & 81: relocated, traveling, transformed, completion, gaining \\
    \texttt{fr} & Chambilly est une commune française, située dans le département de Saône-et-Loire\ldots & 51: tributary, border, flows, passes, alps \\
    \texttt{pt} & Chambilly é uma comuna francesa\ldots no departamento de Saône-et-Loire\ldots & 89: dubbed, estimate, forty, moment, onwards \\
    \texttt{de} & Chambilly ist eine französische Gemeinde\ldots im Département Saône-et-Loire\ldots & 51: tributary, border, flows, passes, alps \\
    \texttt{nl} & Chambilly is een gemeente in het Franse departement Saône-et-Loire\ldots & 21: quebec, nord, maritime, seine, calais\\
    \bottomrule
    \end{tabular}}
    \caption{Sample documents for the topics with highest (top) and lowest (bottom) cross-lingual precision.}
    \label{tab:example}
\end{table*}

\paragraph{Correlation with Existing Benchmarks} To further investigate the role of fine-tuning in inducing better transfer, we employ the Semantic Textual Similarity (STS) benchmark \citep{cer2017semeval};\footnote{This consists of combined English STS data from SemEval shared tasks from 2012--2017. The exact data we use may be downloaded here: \url{https://sbert.net/datasets/stsbenchmark.tsv.gz}} this has been used to evaluate the quality of sentence embeddings more broadly in previous works \citep{sbert,reimers2020multilingsbert}. Performance is evaluated by measuring the Spearman correlation between the cosine similarity of sentence representations and gold labels for the sentence similarity tasks contained in STS. Here, we try correlating this metric with measures of topic quality, as well as with topic transfer (Figure~\ref{fig:sts}). While STS does not correlate strongly with NPMI ($\rho=0.46$, $P>0.1$), it correlates very well with both Match and KL ($\rho=0.93$ and $\rho=0.96$, respectively, and $P<.005$ for both). This implies that {\bf well-tuned sentence embeddings are not necessarily the most important factor in producing good topics, but they are quite important for cross-lingual transfer}. However, cross-lingual transfer performance saturates quickly at STS Spearman coefficients over 55, such that an increase of over 50\% in STS results in only an 8\% increase in Match and 4\% reduction in KL. Thus, one could perhaps trade off STS for better cross-lingual transfer at scores above this threshold. We leave this to future work.

We find further evidence for STS' weak correlation with NPMI and STS' strong correlation with Match and KL when observing the performance of TCCTM: it does not modify the sentence embeddings, so one would expect that TCCTM would perform similarly to the regular CTM if sentence embeddings are of primary importance. This is not the case for NPMI, as TCCTM seems to greatly improve topic quality when using a non-fine-tuned model and have a slightly negative effect when using a fine-tuned model. However, cross-lingual TCCTM performance is consistently comparable to CTM performance with respect to Match and KL when the fine-tuning datasets are the same. 

\subsection{Qualitative Analysis}
Why is fine-tuning important for cross-lingual transfer? Figure~\ref{fig:confmat} displays confusion matrices comparing the topics obtained in English versus those obtained in French for the same documents using both the CTM (not fine-tuned) and CTM+FT (NLI) model. We present confusion matrices for all target languages in Appendix~\ref{app:confmat}. When the embeddings are not fine-tuned, we see that a typical pattern of error is the CTM assigning foreign documents topics from a small subset of the 100 available topics, regardless of the actual content of the document; this is indicated by the frequency of vertical striping in the confusion matrix. After fine-tuning, errors look more evenly distributed across topics and less frequent in general, though there is still slight striping at topic 81. This striping also occurs after fine-tuning at topic 81 for Portuguese and (to a smaller extent) Dutch, but not German. Thus, {\bf CTMs trained on monolingual data are prone to assigning foreign documents topics from a small subset of the available topics, but this can be heavily mitigated with well-tuned sentence embeddings}.

What kinds of topics have high cross-lingual precision, and which have lower precision? We calculate the mean precision per-topic of cross-lingual topic transfer from English to all other target languages using the CTM+FT (NLI) model,\footnote{Recall (and therefore $F_1$) is dominated by topics which are consistently incorrectly assigned to foreign documents---the same topics which cause vertical striping in Figure~\ref{fig:confmat}.} finding that topics which are more qualitatively coherent tend to have higher cross-lingual precision. Topics that are less semantically clear or which compete with similar topics tend to exhibit more cross-lingual variance. Examples of the highest- and lowest-precision topics may be found in Table~\ref{tab:example}.

We sometimes observe competing topics which semantically overlap. In our dataset, this typically occurs for short articles which describe small towns and obscure places, such as in the bottom example of Table~\ref{tab:example}; topics 51 and 21 appear most frequently for these articles. Many instances of topics 81 and 89 (the lowest-precision topics in our dataset) also occur in short articles about small towns or obscure places; we hypothesize that this is often due to the probability mass of more relevant topics being split, thus allowing these topics which contain generally higher-probability tokens to be assigned.

\section{Conclusions}
In monolingual settings, the best topics are achieved through contextualized topic modeling using sentence embeddings fine-tuned on the topic classification task. This holds whether the topic classification objective is used during fine-tuning or integrated into the CTM itself. However, in zero-shot polylingual settings, it is far more important to fine-tune sentence embeddings on any task than to have seen in-domain data during pre-training or to use the topic classification objective. As the topic classification task can be performed on any corpus which has enough documents for topic modeling, supervision for this task is always available; this supervision bootstrapping can therefore serve as a simple way to increase topic quality and transfer for contextualized topic models in the absence of any other data, regardless of domain.

There exists a weak but positive correlation between sentence embedding quality (as measured by the STS benchmark) and topic coherence, but a strong correlation between sentence embedding quality and cross-lingual topic transfer performance. Nonetheless, these preliminary findings also suggest that transfer saturates quickly at quite low STS scores and that STS does not correlate well with topic quality, so we do not necessarily recommend directly optimizing over STS for neural topic modeling.

Future work should investigate fine-tuning on multilingual datasets, as well as explicitly inducing cross-lingual topic alignments. Because the CTM currently generates topics in one language and then transfers into other languages, it would also be beneficial to investigate methods of generating topics in parallel across languages during topic modeling.

\section*{Acknowledgements}
This material is based on work supported by the National Science Foundation Graduate Research Fellowship Program under Grant No.\ 1746891. Any opinions, findings, and conclusions or recommendations expressed in this material are those of the authors and do not necessarily reflect the views of the National Science Foundation.

We wish to thank Shuoyang Ding, Chu-Cheng Lin, and the reviewers for their helpful feedback on earlier drafts of this work.

\bibliography{anthology,naacl2021}
\bibliographystyle{acl_natbib}

\appendix

\clearpage

\section{Confusion Matrices for All Target Languages}\label{app:confmat}

\begin{figure*}
    \centering
    \begin{minipage}{0.49\linewidth}
        \centering
        \includegraphics[width=\linewidth]{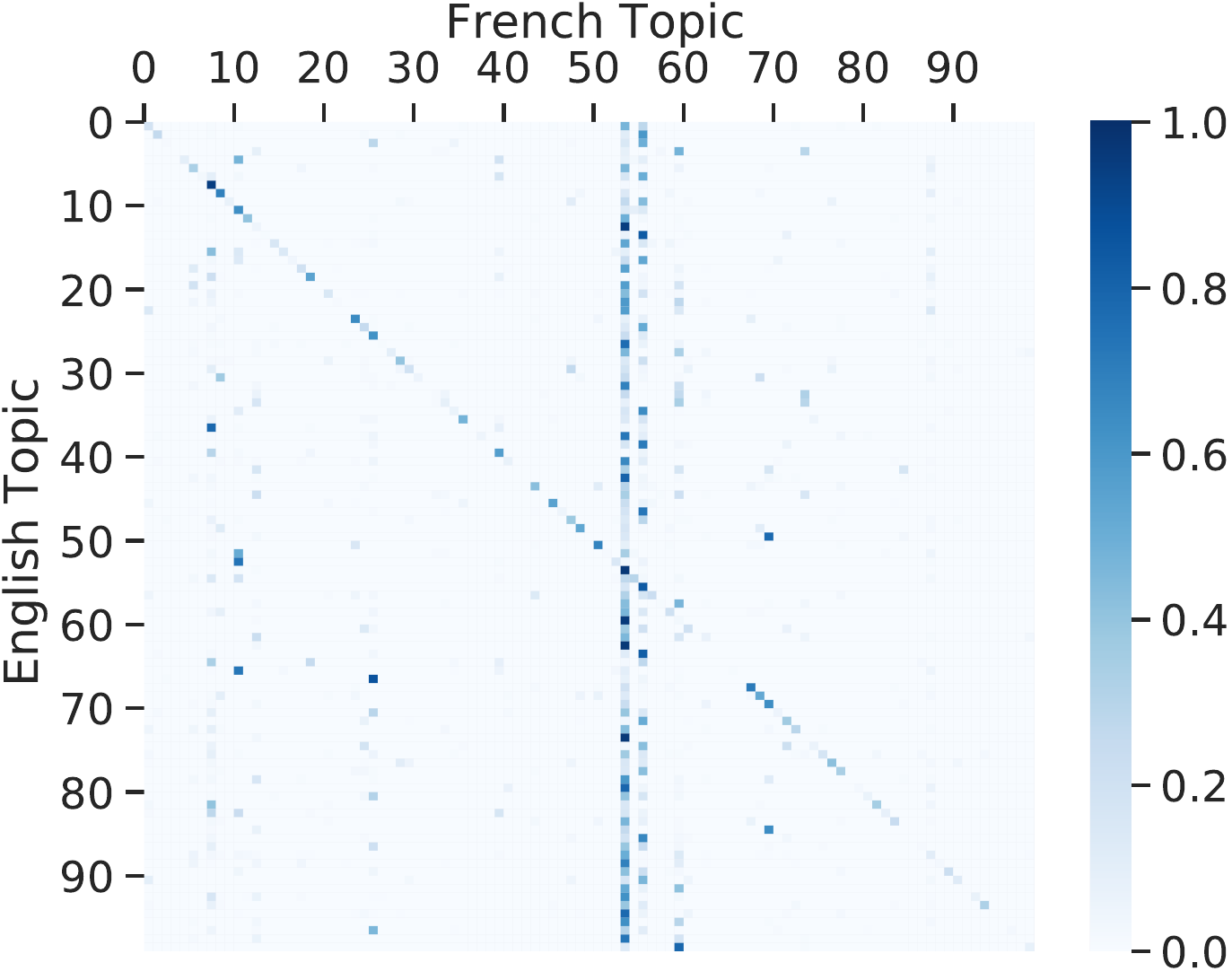}
    \end{minipage}
    \hfill
    \begin{minipage}{0.49\linewidth}
        \centering
        \includegraphics[width=\linewidth]{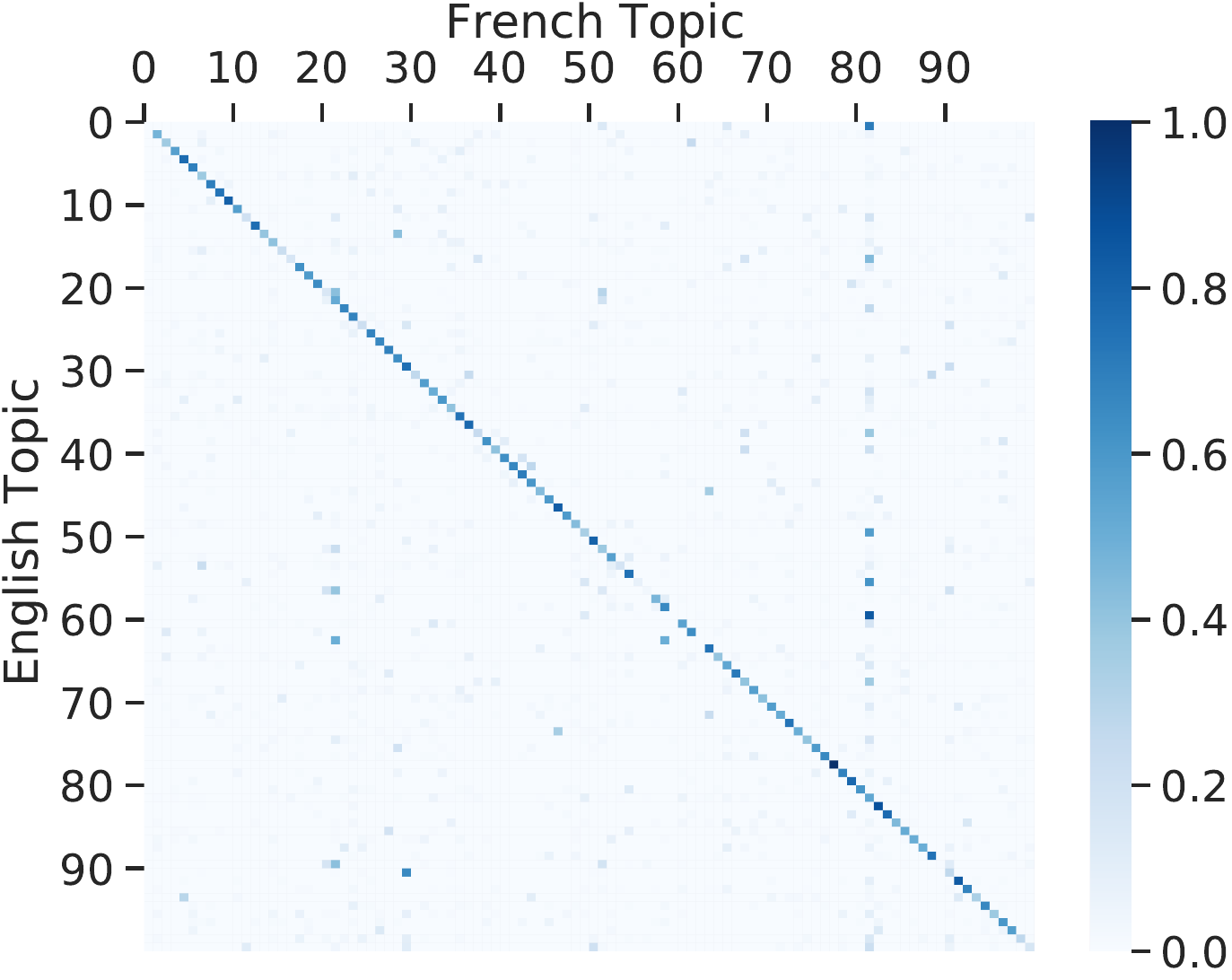}
    \end{minipage}
    \caption{English vs.\ French topic assignments for aligned documents.}
    \label{fig:fr_confmat}
\end{figure*}

\begin{figure*}
    \begin{minipage}{0.49\linewidth}
        \centering
        \includegraphics[width=\linewidth]{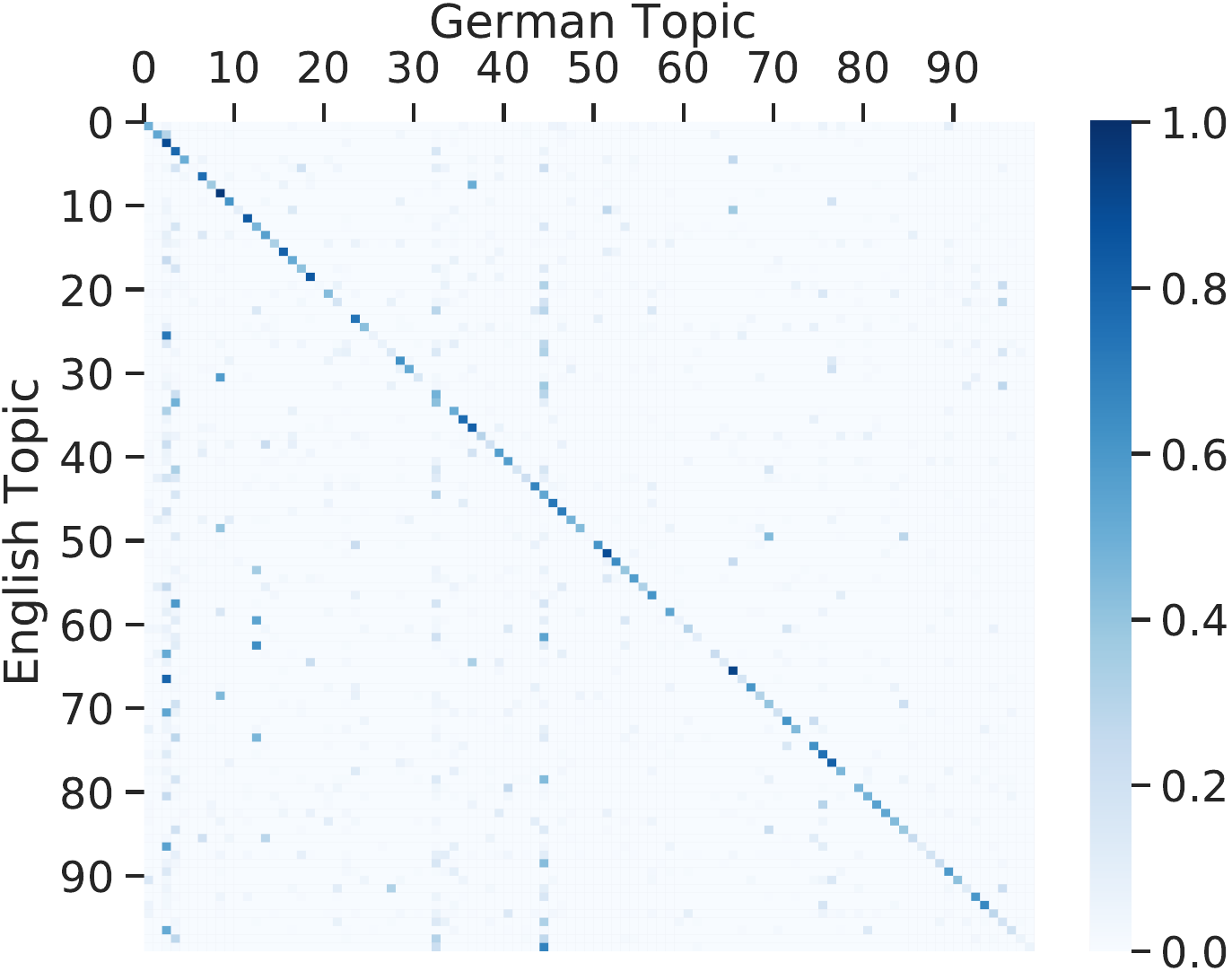}
    \end{minipage}
    \hfill
    \begin{minipage}{0.49\linewidth}
        \centering
        \includegraphics[width=\linewidth]{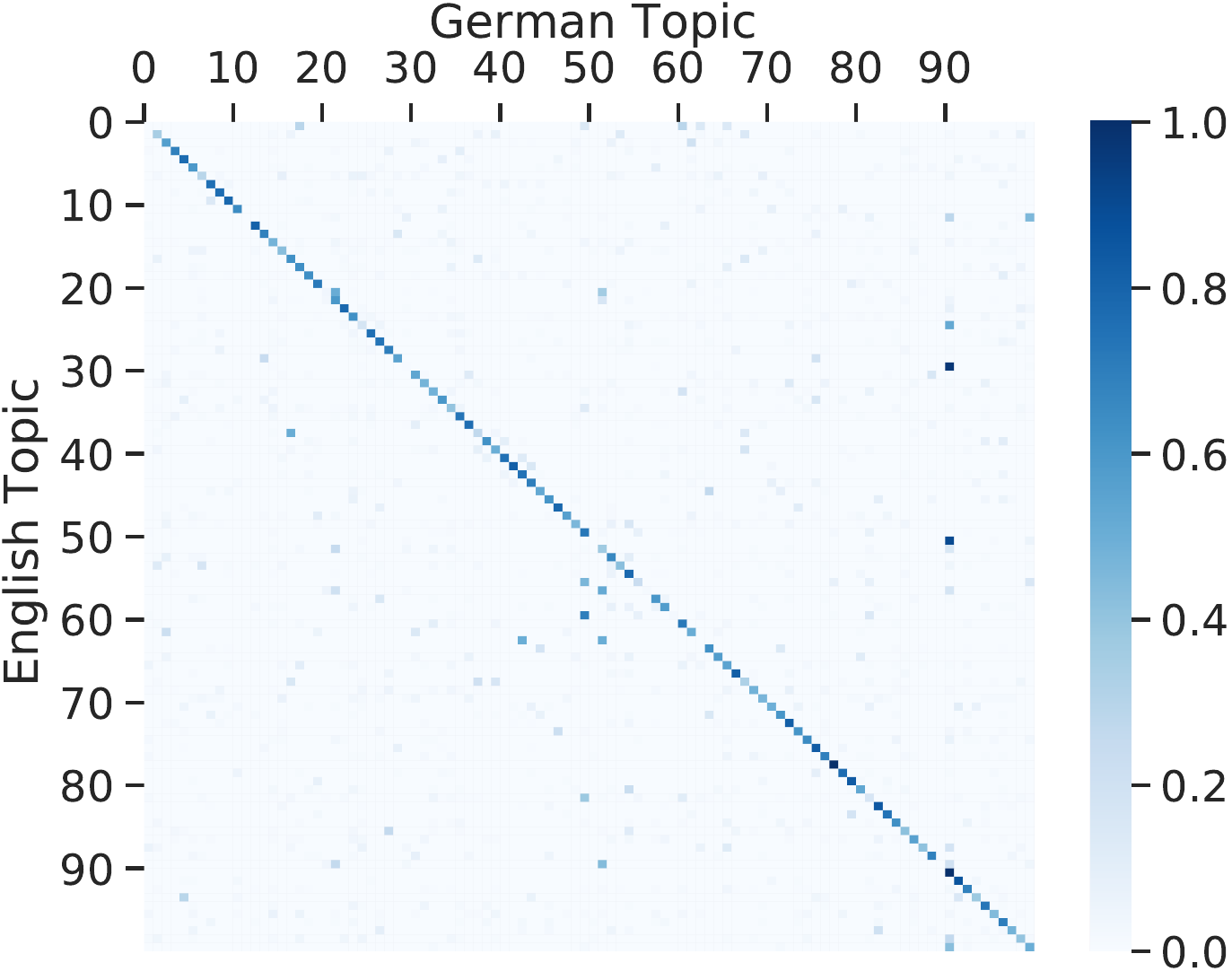}
    \end{minipage}
    \caption{English vs.\ German topic assignments for aligned documents.}
    \label{fig:de_confmat}
\end{figure*}

\begin{figure*}
    \begin{minipage}{0.49\linewidth}
        \centering
        \includegraphics[width=\linewidth]{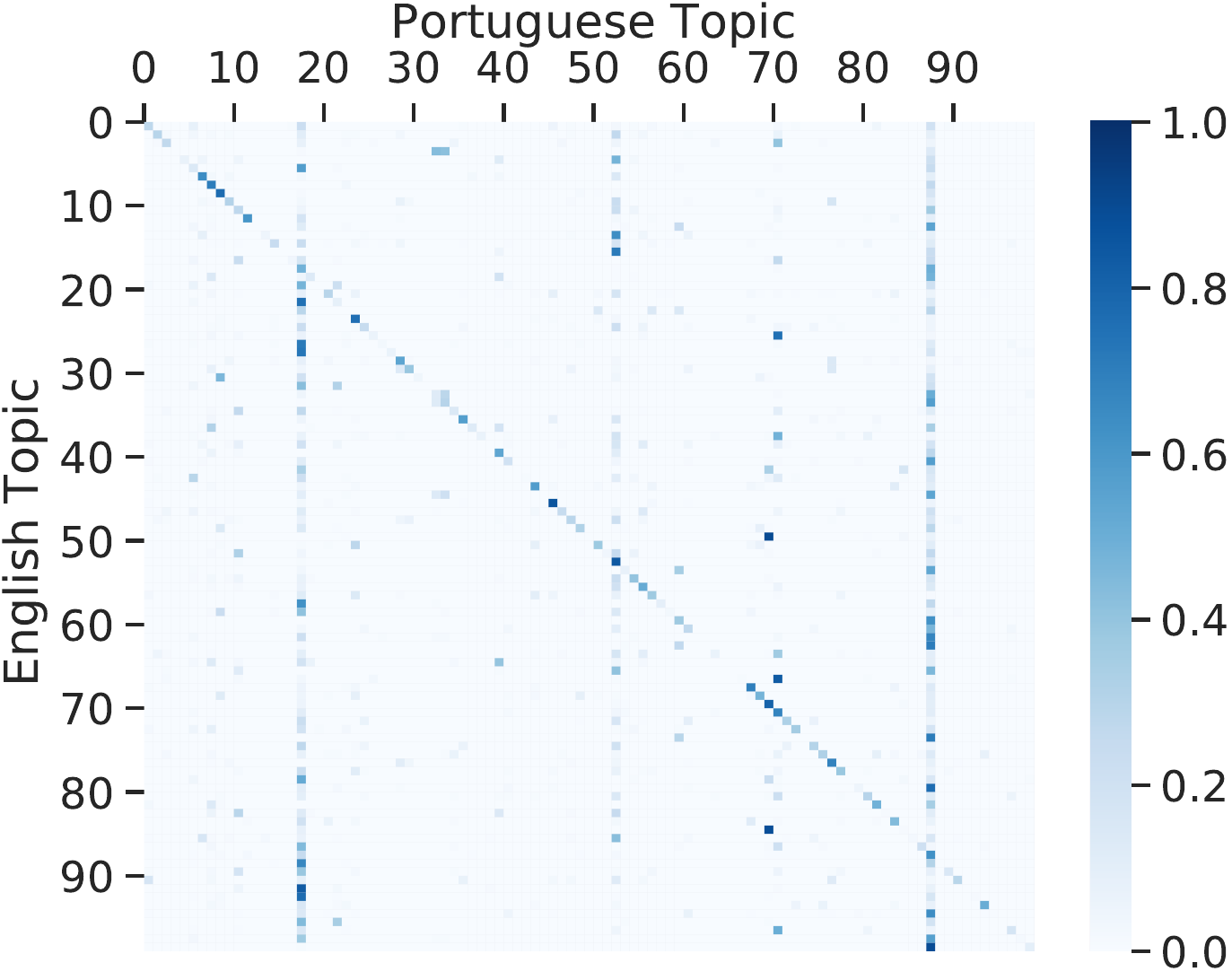}
    \end{minipage}
    \hfill
    \begin{minipage}{0.49\linewidth}
        \centering
        \includegraphics[width=\linewidth]{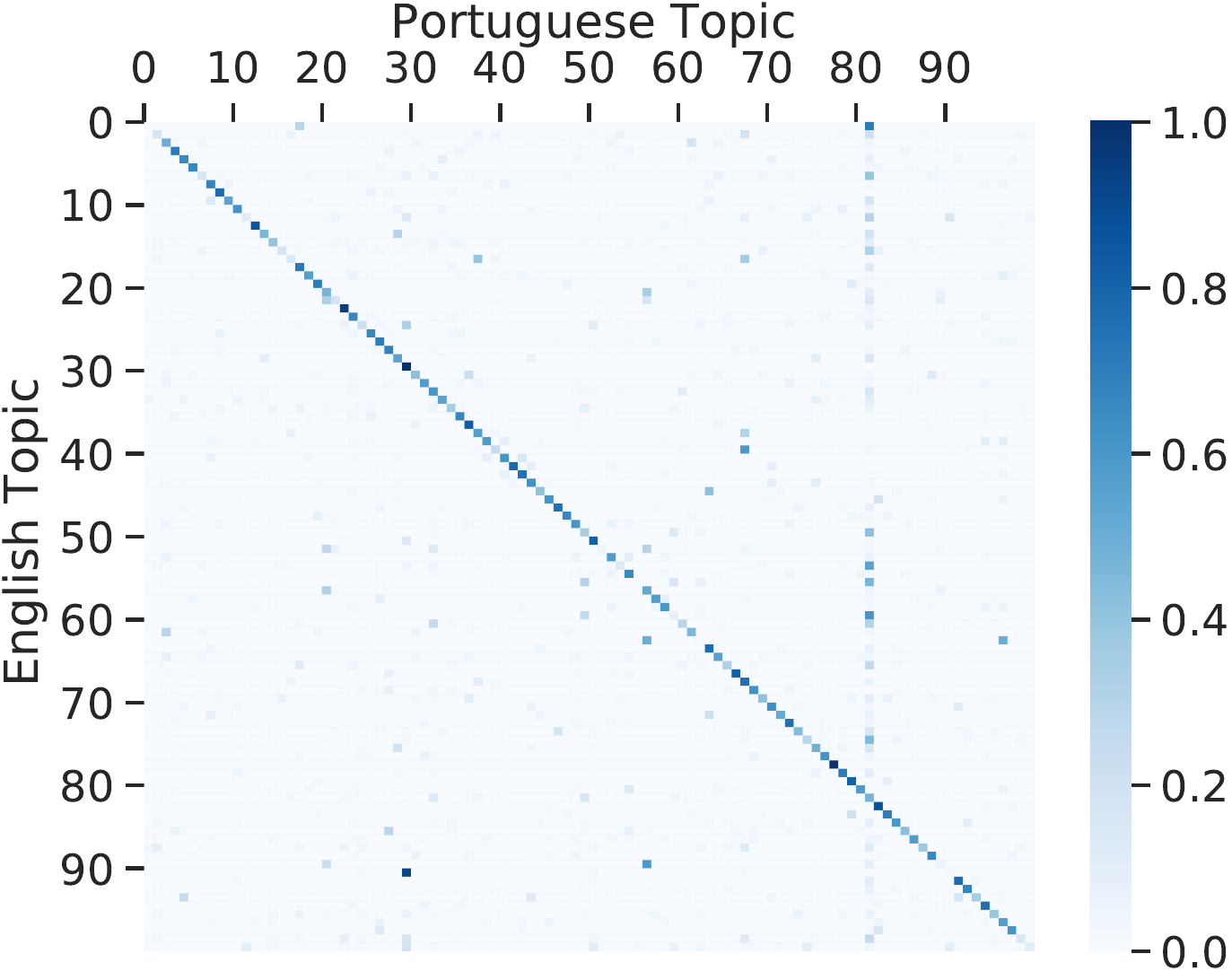}
    \end{minipage}
    \caption{English vs.\ Portuguese topic assignments for aligned documents.}
    \label{fig:pt_confmat}
\end{figure*}

\begin{figure*}
    \begin{minipage}{0.49\linewidth}
        \centering
        \includegraphics[width=\linewidth]{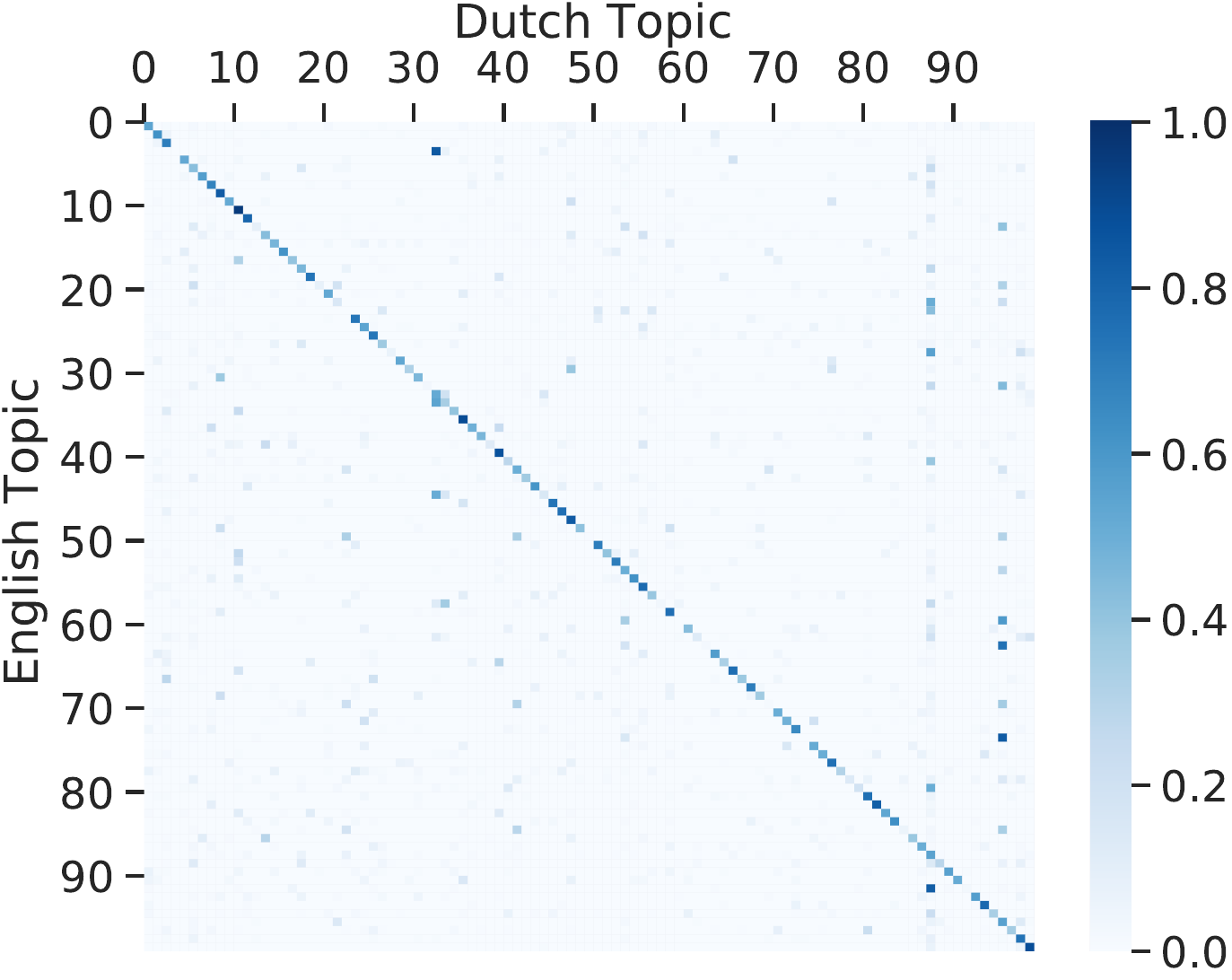}
    \end{minipage}
    \hfill
    \begin{minipage}{0.49\linewidth}
        \centering
        \includegraphics[width=\linewidth]{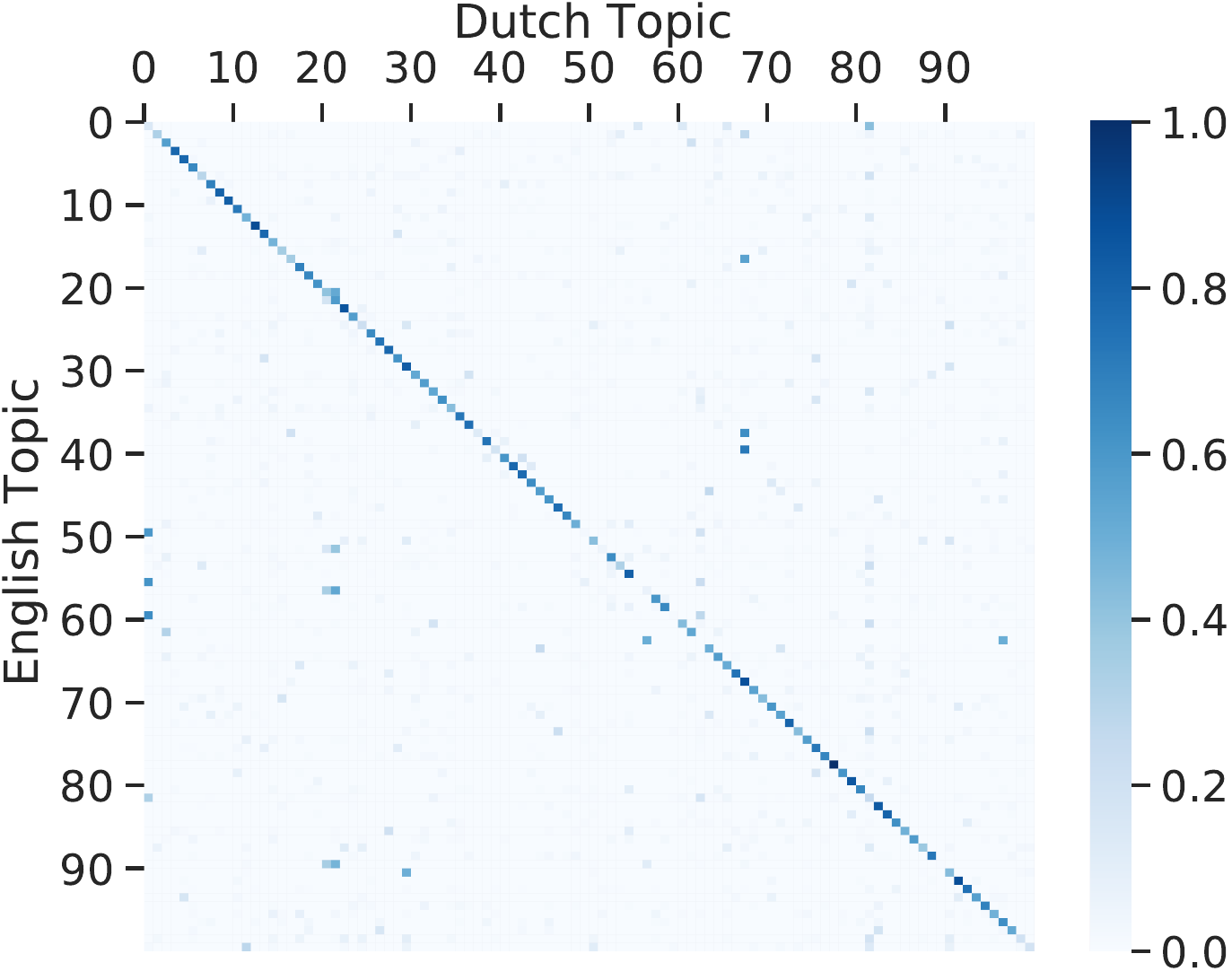}
    \end{minipage}
    \caption{English vs.\ Dutch topic assignments for aligned documents.}
    \label{fig:nl_confmat}
\end{figure*}

\begin{figure*}
    \begin{minipage}{0.49\linewidth}
        \centering
        \includegraphics[width=\linewidth]{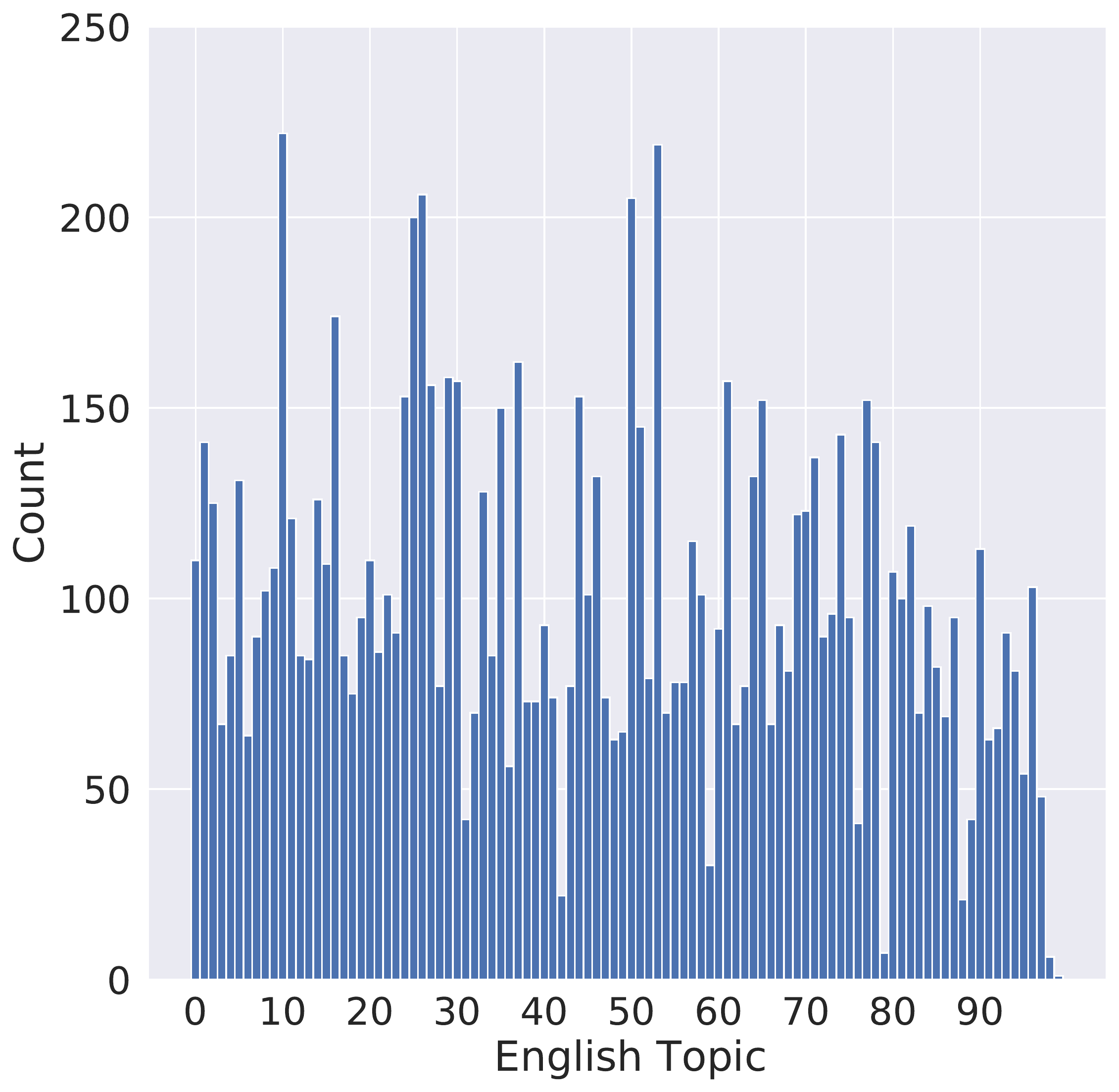}
    \end{minipage}
    \begin{minipage}{0.49\linewidth}
        \centering
        \includegraphics[width=\linewidth]{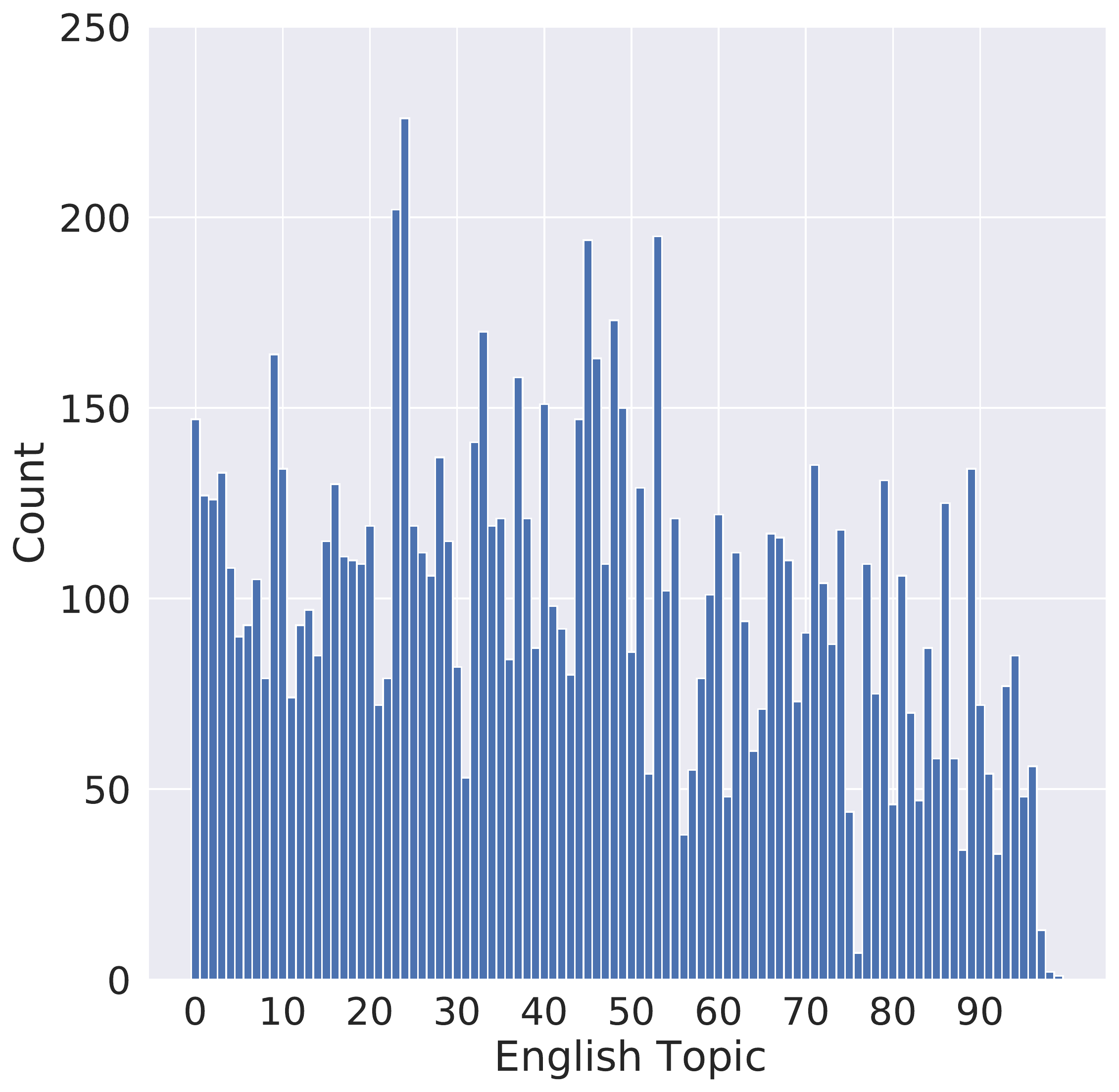}
    \end{minipage}
    \caption{Counts of the most probable topics for each English document in the aligned test set, according to an XLM-R based CTM with non-fine-tuned embeddings (left) and embeddings fine-tuned on NLI (right).}
    \label{fig:counts}
\end{figure*}

Figures~\ref{fig:fr_confmat}, \ref{fig:de_confmat}, \ref{fig:pt_confmat}, and \ref{fig:nl_confmat} present row-normalized confusion matrices comparing topic assignments for aligned documents in English and all other target languages. We present figures for CTMs based on non-fine-tuned embeddings (left) as well as embeddings fine-tuned on NLI (right). All embeddings are based on XLM-R.

\section{English Topic Counts}

As the provided confusion matrices are row-normalized, they do not present the relative frequency of various topics in English. Thus, we present counts of the most probable topics for the English test documents according to a CTM based on non-fine-tuned embeddings and a CTM based on embeddings fine-tuned on NLI (Figure~\ref{fig:counts}).

\section{More Example Topics}\label{app:example}

Table~\ref{tab:example2} presents more sample documents for various high-precision topics. The lowest-precision topics all contain similar top tokens and patterns of error as Table~\ref{tab:example} (as topics 81 and 89 do), so we focus on displaying other types of topics which transfer well across languages.

We find that topics relating to science, sports, places in specific countries, and entertainment transfer well. Perhaps this is due to shared vocabulary for these subjects, as these all contain either scientific terms or proper nouns which are orthographically identical cross-lingually. Or perhaps these subjects are frequently seen during pre-training, thus enabling more isomorphic representations to form around such subjects.

\begin{table*}[h]
    \centering
    \resizebox{\linewidth}{!}{
    \begin{tabular}{lll}
    \toprule
    \bf{Lang} & \bf{Sample Document} & \bf{Topic} \\  
    \midrule
    \texttt{en} & Passiflora, known also as the passion flowers\ldots & 9: cultivated, flower, tall, stems, perennial \\
    \texttt{fr} & Passiflora est un genre de plantes, les passiflores\ldots & 9: cultivated, flower, tall, stems, perennial \\
    \texttt{pt} & Passiflora é um género botânico de cerca\ldots & 9: cultivated, flower, tall, stems, perennial \\
    \texttt{de} & Die art enreiche Pflanzengattung der Passionsblumen\ldots & 7: gene, organisms, cell, biology, dna \\
    \texttt{nl} & Het geslacht passiebloem (Passiflora) is een geslacht\ldots & 9: cultivated, flower, tall, stems, perennial \\
    \midrule
    \texttt{en} & Belgium competed at the 1952 Winter Olympics\ldots & 91: target, boxing, beijing, loser, summer \\
    \texttt{fr} & \ldots la Belgique aux Jeux olympiques d'hiver de 1952\ldots & 91: target, boxing, beijing, loser, summer \\
    \texttt{pt} & A Bélgica competiu nos Jogos Olímpicos de Inverno de 1952\ldots & 91: target, boxing, beijing, loser, summer \\
    \texttt{de} & Belgien nahm an den VI. Olympischen Winterspielen 1952\ldots & 91: target, boxing, beijing, loser, summer \\
    \texttt{nl} & Tijdens de Olympische Winterspelen van 1952\ldots & 91: target, boxing, beijing, loser, summer \\
    \midrule
    \texttt{en} & James Maritato is an American professional wrestler\ldots & 41: wrestler, ring, wwe, heavyweight, professional \\
    \texttt{fr} & \ldots James Maritato\ldots est un ancien catcheur américain\ldots & 41: wrestler, ring, wwe, heavyweight, professional \\
    \texttt{pt} & James Maritato é um lutador de wrestling profissional ítalo-americano\ldots & 41: wrestler, ring, wwe, heavyweight, professional \\
    \texttt{de} & James Maritato\ldots ist ein US-amerikanischer Wrestler\ldots & 41: wrestler, ring, wwe, heavyweight, professional \\
    \texttt{nl} & James Maritato is een Amerikaans professioneel worstelaar\ldots & 41: wrestler, ring, wwe, heavyweight, professional \\
    \midrule
    \texttt{en} & Nemochovice is a village\ldots in the South Moravian Region of the Czech Republic\ldots & 22: czech, prague, bohemian, republic, slovakia \\
    \texttt{fr} & Nemochovice est un village\ldots dans la Moravie-du-Sud en République tchéque\ldots & 22: czech, prague, bohemian, republic, slovakia \\
    \texttt{pt} & Nemochovice é uma comuna checa localizada na região de Morávia do Sul\ldots & 22: czech, prague, bohemian, republic, slovakia \\
    \texttt{de} & Nemochowitz ist eine Gemeinde in Tschechien\ldots & 22: czech, prague, bohemian, republic, slovakia \\
    \texttt{nl} & Nemochovice is een Tsjechische gemeente in de regio Zuid-Moravië\ldots & 22: czech, prague, bohemian, republic, slovakia \\
    \midrule
    \texttt{en} & "Set Fire to the Rain" is a song by British singer-songwriter Adele\ldots & 83: certified, single, billboard, critics, charts \\
    \texttt{fr} & Set Fire to the Rain est une chanson de la chanteuse britannique Adele\ldots & 83: certified, single, billboard, critics, charts \\
    \texttt{pt} & "Set Fire to the Rain" é uma canção da cantora e compositora britânica Adele\ldots & 83: certified, single, billboard, critics, charts \\
    \texttt{de} & Set Fire to the Rain ist ein Lied der britischen Sängerin Adele\ldots & 83: certified, single, billboard, critics, charts \\
    \texttt{nl} & Set Fire to the Rain is de tweede single van Adele's album 21\ldots & 83: certified, single, billboard, critics, charts \\
    \bottomrule
    \end{tabular}}
    \caption{Sample documents for the topics with high cross-lingual precision.}
    \label{tab:example2}
\end{table*}

\end{document}